\documentclass{article}
\usepackage{arxiv}

\usepackage{times}
\usepackage{named}

\usepackage{soul}
\usepackage{url}
\usepackage[hidelinks]{hyperref}
\usepackage[utf8]{inputenc}
\usepackage[small]{caption}
\usepackage{graphicx}
\usepackage{amsmath}
\usepackage{booktabs}
\usepackage{algorithm}
\usepackage{algorithmic}
\usepackage{bm}
\usepackage{amssymb}
\usepackage{amsfonts}
\usepackage{xcolor}
\usepackage{amsthm}
\usepackage{subcaption}
\usepackage{multirow}
\usepackage{comment}
\title{Binary Classification from Positive Data with Skewed Confidence}

\author{
Kazuhiko Shinoda\\
Frontier Research Center, Toyota Motor Corporation\\
\texttt{kazuhiko\_shinoda@mail.toyota.co.jp}\\
\And
Hirotaka Kaji\\
Frontier Research Center, Toyota Motor Corporation\\
\texttt{hirotaka\_kaji@mail.toyota.co.jp}\\
\And
Masashi Sugiyama\\
RIKEN/The University of Tokyo\\
\texttt{sugi@k.u-tokyo.ac.jp}\\
}
\date{\empty}

\begin{document}

\maketitle

\begin{abstract}
\emph{Positive-confidence (Pconf) classification} \cite{Ishida2018} is a promising weakly-supervised learning method which trains a binary classifier only from positive data equipped with confidence.
However, in practice, the confidence may be skewed by bias arising in an annotation process.
The Pconf classifier cannot be properly learned with skewed confidence, and consequently, the classification performance might be deteriorated.
In this paper, we introduce the parameterized model of the skewed confidence, and propose the method for selecting the hyperparameter which cancels out the negative impact of skewed confidence under the assumption that we have the misclassification rate of positive samples as a prior knowledge.
We demonstrate the effectiveness of the proposed method through a synthetic experiment with simple linear models and benchmark problems with neural network models. 
We also apply our method to drivers' drowsiness prediction to show that it works well with a real-world problem where confidence is obtained based on manual annotation.
\end{abstract}

\section{Introduction}

Predicting human behaviour and mental states is a key technology to promote well-being society \cite{Matthews2014,Lane2014}.
Some previous studies trained a predictor by utilizing labeled data collected in laboratory experiments with a relatively small number of subjects \cite{alhanai2017,Shinoda2019}.
On the other hand, large-scale data related to users' physical and physiological activities have been available as IoT devices and smartphones have been widely adopted these days \cite{Weiss2016,Liu2018}.
However, the data collection and annotation processes are still highly expensive when the fully-supervised method is employed.
The cost reduction is therefore a mandatory for the success of real-world applications \cite{Bao2004,Kaji2019}.

For instance, prediction of drowsy driving is one of the important industrial applications \cite{Hachisuka2011}.
A drowsy driving predictor is usually trained with datasets which include both \emph{Alert} and \emph{Drowsy} states collected in experiments using a driving simulator \cite{Esra2007}.
However, driving until drivers feel strong drowsiness is extremely time-consuming because they generally put much effort into staying awake to avoid accidents \cite{horne1999}.
The manual annotation of drivers' states also takes a lot of time and effort since human mental states are ambiguous, and it is hard even for experts to accurately distinguish different states.
Hence, it is quite helpful in saving the cost of data collection and annotation if a drowsiness predictor can be constructed only from \emph{Alert} samples.

\emph{Weakly-supervised classification} \cite{Zhou2018} is an essential approach to drastically reduce the annotation costs.
Various problem formulations have been considered so far depending on what types of information is available as weakly-labeled data, which are less informative but less expensive than fully-labeled data \cite{Mintz2009,Bao2018}.
\emph{Semi-supervised classification} \cite{Chapelle2010,Zhu2003,Chapelle2005,Sakai2017}, positive-unlabeled (PU) classification \cite{Elkan2008,duPlessis2014}, and unlabeled-unlabeled classification \cite{Lu2019,duPlessis2013} are the typical examples of weakly-supervised classification to utilize unlabeled data.
There have also been studies in the other problem settings including multiple instance classification \cite{Dietterich1997}, partial label classification \cite{Cour2011}, complementary classification \cite{Ishida2017}, and similar-unlabeled classification \cite{Bao2018}.
In addition, noisy-label classification \cite{Liu2016,Natarajan2013} and PU classification with selection bias \cite{Kato2019} are methods for dealing with uncertainties in classification, which is an important research topic from a practical viewpoint.

In this paper, we pay special attention to positive-confidence (Pconf) classification \cite{Ishida2018}, which is a unique weakly-supervised binary classification method that can learn only from positive data and the confidence.
Compared with anomaly detection methods such as the one-class support vector machine \cite{Scholkopf2001}, Pconf classification can practically achieve high classification accuracy and has a theoretical guarantee in the framework of empirical risk minimization.
However, the performance of the Pconf classifier may be deteriorated when the confidence is skewed.
We will use the term \emph{skew} henceforth in this paper to indicate that the distribution of the positive confidence is deformed due to bias, which often occurs in an annotation process.

In practice, the confidence of positive samples may be obtained based on a tacit knowledge learned from past experience.
The confidence can easily be skewed in this situation when there is bias caused by a lack of experience (the sample size experienced in the past was small) or imbalanced experience (abundant knowledge on a positive class but little knowledge on a negative class) for example.

\cite{Ishida2018} has demonstrated that the Pconf classification was robust against noisy confidence.
However, to the best of our knowledge, the more general case of skewed confidence has not been considered for Pconf classification.
The purpose of this paper is to extend Pconf classification to be able to cope with skewed confidence, which will significantly enhance its practicality.

The challenge of this paper is to give a practical method to address the problem of skewed confidence in the Pconf classification setting.
In order to correct the skewed confidence, we introduce a parameterized model of the confidence.
However, we may not be able to directly optimize the confidence model since we only have positive samples.
Our key idea to overcome this difficulty is to assume a prior knowledge, the misclassification rate of positive samples, 
and optimize the confidence model through the minimization of the squared difference between the misclassification rate and empirical validation error.
Is this assumption reasonable?
As typified by the examples below, some real-world problems fit well with our assumption:
\begin{description}
\item[Example 1] Store managers want to predict whether their customers will continue to come to their stores or not based on the customers' attributes data and loyalty score.
They cannot employ fully-supervised methods since they do not have data of rival stores while Pconf classification is available by using the loyalty score as the positive confidence.
However, the evaluation of the loyalty score tends to be too optimistic because the store managers do not have the information on non-customers.
On the other hand, they empirically know the overall churn rate.
\item[Example 2] Engineers want to predict whether a person is sleepy or not from her physiological information.
The self-report score of drowsiness can be used as the ground truth \cite{Shinoda2019} to train a Pconf classifier.
They empirically know the ratio of drowsiness in daytime, but the person may underreport the score because of embarrassment when she feels the strong drowsiness.
\end{description}
Then we experimentally demonstrate the effectiveness of the proposed method through synthetic toy problems with linear-in-input models and benchmark problems with neural network models. 
Finally, we apply our method to a real-world problem of drivers' drowsiness prediction to show the practical usefulness.

\section{Problem Formulation}

In this section, we review the original problem setting of Pconf classification \cite{Ishida2018}.

Let $\bm{x}\in\mathbb{R}^d$ be a $d$-dimensional pattern and $y\in\{+1,-1\}$ be its class label.
Consider a situation where positive samples and the confidence are only available for training a classifier:
\[
\mathcal{X} := \{(\bm{x}_{i}, r_{i}) \}_{i=1}^{n}, 
\]
where $\bm{x}_{i}$ is a positive sample independently drawn from $p(\bm{x}|y=+1)$, 
and $r_{i}=p(y=+1|\bm{x}_i)$ is the confidence.
Because we do not have negative samples under this problem setting, 
the classification risk minimization represented by the following formulation cannot be directly executed:
\[
R(g) = \mathbb{E}_{p(\bm{x},y)}[\ell(yg(\bm{x}))],
\]
where $p(x,y)$ is the joint density that test data follows, $g(\bm{x}) : \mathbb{R}^{d} \rightarrow \mathbb{R}$ is a binary classifier, and $\ell(z)$ is a loss function.
Let $\pi_{+}=p(y=+1)$ and $r(\bm{x})=p(y=+1|\bm{x})$, and let $\mathbb{E}_{+}$ denote the expectation over $p(\bm{x}|y=+1)$.
Then the classification risk $R(g)$ is written as follows \cite{Ishida2018}:
\[
R(g) = \pi_{+}\mathbb{E}_{+} \left[ \ell\left(g(\bm{x})\right) + \frac{1-r(\bm{x})}{r(\bm{x})} \ell\left(-g(\bm{x})\right) \right].
\]
When minimizing $R(g)$ with respect to $g$, $\pi_{+}$ can be neglected since it is a proportional constant.
Therefore, empirical risk minimization can be executed by using positive samples and the confidence as follows:
\[
\min_{g} \sum_{i=1}^{n} \left[ \ell\left(g(\bm{x}_{i})\right) + \frac{1-r_{i}}{r_{i}} \ell\left(-g(\bm{x}_{i})\right) \right].
\]

\section{Adjusted Pconf Classification}

We then introduce our problem setting, that is, Pconf classification with skewed confidence, and propose the method for alleviating the negative impact of skew.

In order to correct skewed confidence $\hat{r}$, we employ the following formulation with adjusted confidence $r_i^k$ for $0<k<\infty$:
\[
\min_{g} \sum_{i=1}^{n} \left[ \ell\left(g(\bm{x}_{i})\right) + \frac{1-r_{i}^{k}}{r_{i}^{k}} \ell\left(-g(\bm{x}_{i})\right) \right].
\]
Note that $r_i^{k} \in (0, 1]$ and $r_{i} \leq r_{j} \Leftrightarrow r_{i}^{k} \leq r_{j}^{k}$.
In this model, the skewed confidence is supposed as $\hat{r} = r^{-k}$
\footnote{In this paper, we consider the exponential model for the ease of controlling the flatness of confidence distribution $r(\bm{x})$.
Other adjusted confidence models such as an additive model $\max \left(0, \min(1, r+k ) \right), -1 \leq k \leq 1$ can also be employed.
}.

The hyperparameter $k$ may be selected via cross validation if we have a validation set which includes both positive and negative samples.
However, we cannot optimize $k$ since we only have positive samples in the current setup.
Thus it may not be possible to figure out the presence of skew in confidence and its magnitude only from positive samples.

To overcome this difficulty, we assume that we know the misclassification rate of positive samples $\phi$ as a prior knowledge:
\[
\phi = \int_{\{\bm{x}:g(\bm{x})<0\}} p(\bm{x}|y=+1)d\bm{x}.
\]
Under this assumption, optimal hyper-parameter $k^{*}$ may be selected by minimizing the squared error between the empirical classification error and $\phi$:
\[
k^{*} = \arg \min_{k} \left( \frac{1}{n}\sum_{i=1}^{n}\ell_{01}\left(g(\bm{x}_{i})\right)-\phi \right)^{2},
\]
where $\ell_{01}(z) = (1-\mathrm{sign}(z))/2$ is the zero-one loss.
Finally, the adjusted Pconf classifier is obtained via the following optimization problem:
\[
\min_{g} \sum_{i=1}^{n} \left[ \ell\left(g(\bm{x}_{i})\right) + \frac{1-r_{i}^{k^{*}}}{r_{i}^{k^{*}}} \ell\left(-g(\bm{x}_{i})\right) \right].
\]

\section{Experiments}
We examine the behaviour of our proposed method on simple synthetic data, and demonstrate the performance of our method on benchmark image datasets.
We implemented the experiments on Python using Optuna \cite{optuna}, PyTorch \footnote{\url{https://pytorch.org/}}, and Scikit-learn \cite{sklearn}.

\subsection{Synthetic Experiments}
In this toy experiment, a training set, a validation set, a test set, and a dataset for estimating confidence were created from positive and negative samples drawn independently from two-dimensional Gaussian distributions.
All the dataset except the validation set consisted of 1,000 positive samples and 1,000 negative samples.
The validation set included only 1,000 positive samples.
We computed the probabilistic output of $\ell_{2}$-regularized linear logistic regression as positive confidence, and transforming it to skewed confidence by taking the $b$-th power ($b\in\left\{0.3,0.5,2.0,4.0\right\}$).
Confidence lower than 0.01 was rounded up to 0.01 to stabilize the optimization.

We compared original Pconf, adjusted Pconf, and the fully-supervised method.
Linear-in-input model $g(\bm{x})=\bm{\alpha}^{\top}\bm{x}+\beta$ and the logistic loss $\ell_{L}\left(z\right)=\log\left(1+e^{-z}\right)$ were used for all of these methods. Then the empirical risk minimization for Pconf classification can be formulated as follows:
\[
\min_{\bm{\alpha}} \sum_{i=1}^{n} \left[ \log\left(1+e^{-\bm{\alpha}^{\top}\bm{x}_{i}-\beta}\right) + \frac{1-r_{i}}{r_{i}} \log\left(1+e^{\bm{\alpha}^{\top}\bm{x}_{i}+\beta}\right) \right].
\]
\emph{Adam} \cite{adam} with 5,000 epochs was used for optimization.
Note that negative samples were not used for training the original Pconf and adjusted Pconf classifiers, and selecting the hyperparameter $k$.

\begingroup
\renewcommand{\arraystretch}{1.3}
\begin{table}[!t]
\centering
\caption{The mean and standard deviation of the classification accuracy over 10 trials with different degrees of class overlap and skewed confidence. Adjusted Pconf classification was compared with original Pconf classification and fully-supervised classification. Bold face denotes the best and comparable methods according to the paired $t$-test at the significance level of 5\% between original Pconf and adjusted Pconf. We also report the mean and standard deviation of the false negative rate over 10 trials with fully-supervised classification as $\widehat{\phi}$.}
\begin{tabular}{cc|cc|c|c} \hline
$\bm{\mu}_{-}$ & $b$ & O. Pconf & A. Pconf & Supervised & $\widehat{\phi}$ \\ \hline
\multirow{4}{*}{1.0} & 0.3 & 60.73 $\pm$ 0.81 & \textbf{75.78} $\pm$ \textbf{0.77} & \multirow{4}{*}{75.90 $\pm$ 0.74} & \multirow{4}{*}{23.47 $\pm$ 1.11} \\
 & 0.5  & 69.52 $\pm$ 1.18 & \textbf{75.81} $\pm$ \textbf{0.74} &  \\
 & 2.0  & 71.05 $\pm$ 0.73 & \textbf{75.78} $\pm$ \textbf{0.89} &  \\
 & 4.0  & 63.57 $\pm$ 0.80 & \textbf{75.81} $\pm$ \textbf{0.73} &  \\ \hline
\multirow{4}{*}{1.5} & 0.3 & 75.09 $\pm$ 0.73 & \textbf{85.74} $\pm$ \textbf{0.87} & \multirow{4}{*}{85.82 $\pm$ 0.98} & \multirow{4}{*}{14.36 $\pm$ 0.93} \\
 & 0.5  & 81.73 $\pm$ 0.68 & \textbf{85.76} $\pm$ \textbf{0.89} &  \\
 & 2.0  & 82.97 $\pm$ 1.22 & \textbf{85.81} $\pm$ \textbf{0.83} &  \\
 & 4.0  & 77.46 $\pm$ 1.27 & \textbf{85.59} $\pm$ \textbf{0.61} &  \\ \hline
\multirow{4}{*}{2.0} & 0.3 & 85.97 $\pm$ 1.21 & \textbf{92.02} $\pm$ \textbf{0.40} & \multirow{4}{*}{92.05 $\pm$ 0.36} & \multirow{4}{*}{7.85 $\pm$ 0.99} \\
 & 0.5  & 90.02 $\pm$ 0.83 & \textbf{92.01} $\pm$ \textbf{0.43} &  \\
 & 2.0  & 90.49 $\pm$ 0.60 & \textbf{91.95} $\pm$ \textbf{0.44} &  \\
 & 4.0  & 87.14 $\pm$ 0.53 & \textbf{92.04} $\pm$ \textbf{0.38} &  \\ \hline
\multirow{4}{*}{2.5} & 0.3 & 92.89 $\pm$ 0.74 & \textbf{96.27} $\pm$ \textbf{0.57} & \multirow{4}{*}{96.24 $\pm$ 0.52} & \multirow{4}{*}{3.95 $\pm$ 0.53} \\
 & 0.5  & 95.32 $\pm$ 0.45 & \textbf{96.24} $\pm$ \textbf{0.55} &  \\
 & 2.0  & 95.17 $\pm$ 0.56 & \textbf{96.25} $\pm$ \textbf{0.56} &  \\
 & 4.0  & 93.22 $\pm$ 0.68 & \textbf{96.25} $\pm$ \textbf{0.55} &  \\ \hline
\multirow{4}{*}{3.0} & 0.3 & 96.42 $\pm$ 0.66 & \textbf{98.15} $\pm$ \textbf{0.38} & \multirow{4}{*}{98.14 $\pm$ 0.30} & \multirow{4}{*}{1.72 $\pm$ 0.44} \\
 & 0.5  & 97.73 $\pm$ 0.39 & \textbf{98.14} $\pm$ \textbf{0.35} &  \\
 & 2.0  & 97.57 $\pm$ 0.48 & \textbf{98.13} $\pm$ \textbf{0.36} &  \\
 & 4.0  & 96.23 $\pm$ 0.62 & \textbf{98.13} $\pm$ \textbf{0.44} &  \\ \hline
\end{tabular}
\label{syn1}
\end{table}
\endgroup

\begin{figure}[!ht]
  \centering
  \includegraphics[width=0.59\linewidth]{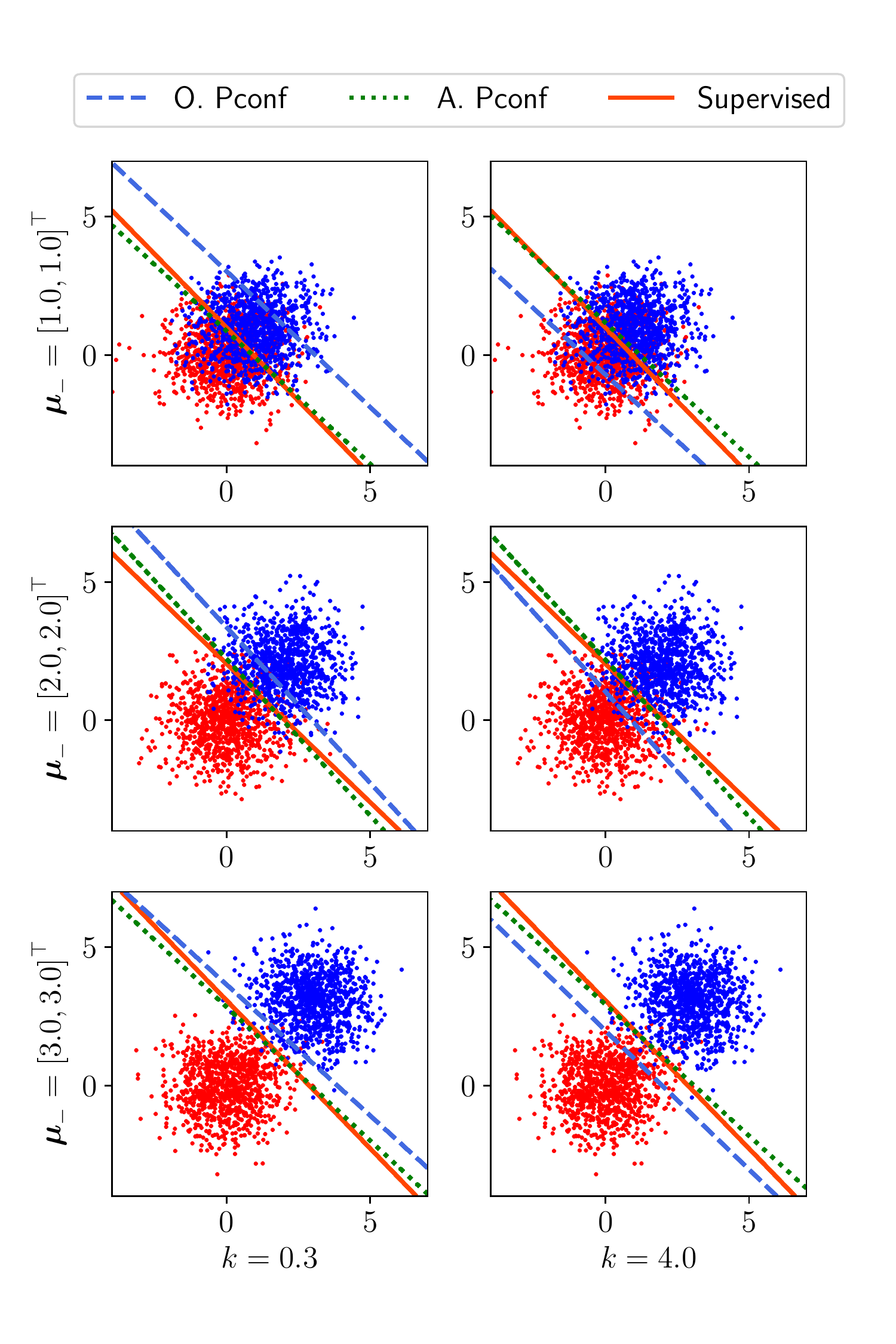}
  \caption{The visual illustrations of the impact of the different degrees of class overlap and skewed confidence. The red and blue points represent the positive and negative samples, respectively. The decision boundaries of original Pconf, adjusted Pconf and fully-supervised classification are shown as the blue, green and red lines, respectively.}
  \label{synfig}
\end{figure}

\subsubsection{Impact of Degree of Class Overlap and Skewed Confidence}
We first examine the impact of the degree of class overlap and skewed confidence on the behaviour of Pconf classification.
We fixed the mean of the positive distribution $\bm{\mu}_{+}$ to $\left[0,0\right]^{\top}$, and moved the mean of the negative distribution $\bm{\mu}_{-}$ from $\left[1.0,1.0\right]^{\top}$ to $\left[3.0,3.0\right]^{\top}$.
We used the identity covariance for both the positive and negative distributions.
In addition, we changed the degree of skewed confidence $b\in\left\{0.3,0.5,2.0,4.0\right\}$.
We calculated the mean false negative (FN) rate over 10 trials with fully-supervised classification as $\widehat{\phi}$ (an estimate of $\phi$).

The mean and standard deviation of the classification accuracy over 10 trials are reported in Table \ref{syn1}.
Adjusted Pconf worked well with all the candidates of the degree of class overlap and skewed confidence.
It was statistically significantly better than original Pconf in all the cases, but difference in accuracy got smaller as the negative distribution moved away from the positive distribution.
This result indicates that skewed confidence has less impact on problems where two distributions are easily separable because a small change of a decision boundary does not affect the classification result in such a situation.
The visual illustrations of the results are presented in Figure \ref{synfig}.

\subsubsection{Impact of Estimation Error of $\phi$}
In the first experiment, we assumed that we had access to a precise estimate of the misclassification rate of positive samples $\phi$.
However, this assumption would not always hold in practice.
We then conducted the experiments with different values of $\widehat{\phi}$ to see the robustness of our proposed method against the estimation error of $\phi$.
We used the mean FN rate reported in Table \ref{syn1} multiplied by $c\in\left\{0.5,0.7,1.3,1.5\right\}$ as $\widehat{\phi}$ with estimation error.

Table \ref{syn2} presents the results where $b=0.3$ and $4.0$.
Although the classification accuracy decreased compared to the case with no estimation error in $\phi$, the decrease in accuracy was insignificant in some cases.
Additionally, the accuracy of adjusted Pconf in Table \ref{syn2} was better than the accuracy of original Pconf reported in Table \ref{syn1} even for the cases with 50\% estimation error ($c=0.5$ and $1.5$).
This result demonstrates the usefulness of our proposed method in a practical situation where the value of $\phi$ cannot be precisely estimated.

\begingroup
\renewcommand{\arraystretch}{1.3}
\begin{table}[!t]
\centering
\caption{The mean and standard deviation of the classification accuracy over 10 trials using $\widehat{\phi}$ with estimation error. Bold face shows that the result is comparable to the case where $c=1.0$ (no error) according to the paired $t$-test at the significance level of 5\%.}
\begin{tabular}{cc|cccc} \hline
$b$ & $\bm{\mu}_{-}$ & $c=0.5$ &  $c=0.7$ &  $c=1.3$ &  $c=1.5$ \\ \hline
\multirow{5}{*}{0.3} & 1.0 & 73.08 $\pm$ 0.98  & 74.94 $\pm$ 0.87 & \textbf{75.30} $\pm$ \textbf{0.93} & 74.94 $\pm$ 0.64 \\
& 1.5 & 83.44 $\pm$ 0.95  & 84.72 $\pm$ 1.03 & \textbf{85.22} $\pm$ \textbf{0.77} & \textbf{84.75} $\pm$ \textbf{1.03} \\
& 2.0 & 90.81 $\pm$ 0.94  & 91.78 $\pm$ 0.71 & \textbf{91.70} $\pm$ \textbf{0.50} & 91.46 $\pm$ 0.52 \\
& 2.5 & 95.30 $\pm$ 0.99  & 95.80 $\pm$ 0.60 & 95.90 $\pm$ 0.42 & 95.74 $\pm$ 0.48 \\
& 3.0 & \textbf{98.06} $\pm$ \textbf{0.60}  & \textbf{98.21} $\pm$ \textbf{0.56} & \textbf{98.30} $\pm$ \textbf{0.25} & \textbf{98.19} $\pm$ \textbf{0.19} \\ \hline
\multirow{5}{*}{4.0} & 1.0 & 73.84 $\pm$ 0.66  & \textbf{75.39} $\pm$ \textbf{0.58} & 75.40 $\pm$ 0.69 & 74.9 $\pm$ 0.76 \\
& 1.5 & 83.07 $\pm$ 0.58  & 84.55 $\pm$ 0.62 & \textbf{84.96} $\pm$ \textbf{1.00} & 84.60 $\pm$ 1.02 \\
& 2.0 & 91.32 $\pm$ 0.67  & 91.86 $\pm$ 0.56 & \textbf{92.08} $\pm$ \textbf{0.49} & 91.72 $\pm$ 0.44 \\
& 2.5 & 95.73 $\pm$ 0.43  & \textbf{96.15} $\pm$ \textbf{0.30} & \textbf{96.14} $\pm$ \textbf{0.47} & 95.88 $\pm$ 0.55 \\
& 3.0 & \textbf{97.92} $\pm$ \textbf{0.54}  & \textbf{98.17} $\pm$ \textbf{0.34} & \textbf{98.27} $\pm$ \textbf{0.42} & \textbf{98.19} $\pm$ \textbf{0.43} \\ \hline
\end{tabular}
\label{syn2}
\end{table}
\endgroup

\subsection{Benchmark Experiments}
We conducted two benchmark experiments with deep neural network models to evaluate the performance of our proposed method on image datasets.
The experimental setups were based on \cite{Ishida2018}.

\subsubsection{Fashion-MNIST}
The Fashion-MNIST dataset is a set of 28 $\times$ 28 gray-scale images each of which represents one of the following 10 fashion item classes: T-shirt/top, pullover, dress, coat, sandal, shirt, sneaker, bag, and ankle boot. Each class consists of 7,000 images.
We chose T-shirt/top as the positive class and one of the others as the negative class to construct a dataset for binary classification.
We then divided it into four stratified sub-datasets: a training set, a validation set, a test set, and a dataset for estimating confidence.

We compared the performance of original Pconf classification, adjusted Pconf classification and fully-supervised classification.
Skewed confidence was used for training both original and adjusted Pconf classifiers.
We also included in this experiment original Pconf classification with non-skewed confidence to evaluate the performance decline caused by skew in confidence.
The logistic loss was used for all of these methods.

We used a three-layer fully-connected neural network (784-100-100-1) with ReLU \cite{relu} as an activation function and $10^{-7}$ weight decay.
Optimization was executed by \emph{Adam} \cite{adam} for 200 epochs with minibatch size 100.

We generated positive confidence from a neural network with a softmax output layer and the same architecture as the aforementioned binary classifiers, but we used 50\% dropout \cite{dropout} after each fully-connected layer instead of weight decay.
Additionally, we set the number of epochs to 20 in order to prevent the probabilistic output from being too close to 0 or 1, which is not suitable for the representation of confidence.
We made confidence skewed by removing 90\% negative samples from a dataset for training a softmax network.
As a result, the dataset became imbalanced (P : N = 10 : 1), and confidence was skewed towards the positive class.
Confidence lower than 0.01 was rounded up to 0.01 to stabilize the optimization.
We used the mean FN rate over 20 trials with fully-supervised classification as an estimate of $\phi$.

The mean and standard deviation of the classification accuracy over 20 trials are reported in Table \ref{fashion}.
Although skewed confidence severely affected the performance of original Pconf, it performed well with the accuracy over 90\% in some cases where the supervised method had relatively high accuracy.
This result was consistent with the synthetic experiment.
Adjusted Pconf outperformed original Pconf in most of the cases, and was even competitive with fully-supervised classification and original Pconf with non-skewed confidence in some cases.

\begingroup
\renewcommand{\arraystretch}{1.3}
\begin{table}[!t]
\centering
\caption{The mean and standard deviation of the classification accuracy over 20 trials using three-layer fully-connected neural network models on the Fashion-MNIST dataset. Adjusted Pconf classification was compared with original Pconf classification, original Pconf classification with non-skewed confidence, and fully-supervised classification. We used the T-shirt/top as the positive class in all cases. Bold face denotes the best and comparable methods according to the paired $t$-test at the significance level of 5\% between original Pconf and adjusted Pconf.}
\begin{tabular}{l|ll|l|l} \hline
Negative & \multicolumn{1}{c}{O. Pconf} & \multicolumn{1}{c}{A. Pconf} & \multicolumn{1}{|c}{Non-skewed} & \multicolumn{1}{|c}{Supervised}  \\ \hline
trouser & 65.30 $\pm$ 9.92  & \textbf{91.72} $\pm$ \textbf{1.85} & 95.64 $\pm$ 0.30 & 99.12 $\pm$ 0.13 \\
pullover & 50.00 $\pm$ 0.00  & \textbf{84.31} $\pm$ \textbf{2.04} & 96.31 $\pm$ 0.12 & 96.90 $\pm$ 0.29 \\
dress & 50.00 $\pm$ 0.00  & \textbf{72.27} $\pm$ \textbf{2.79} & 91.13 $\pm$ 0.25 & 95.81 $\pm$ 0.55 \\
coat & 50.02 $\pm$ 0.07  & \textbf{85.42} $\pm$ \textbf{1.70} & 97.02 $\pm$ 0.19 & 98.61 $\pm$ 0.26 \\
sandal & \textbf{93.76} $\pm$ \textbf{10.41}  & \textbf{96.42} $\pm$ \textbf{2.28} & 98.02 $\pm$ 0.84 & 99.84 $\pm$ 0.04 \\
shirt & 50.00 $\pm$ 0.00  & \textbf{70.12} $\pm$ \textbf{2.88} & 81.76 $\pm$ 0.14 & 85.21 $\pm$ 0.70 \\
sneaker & 97.66 $\pm$ 2.04  & \textbf{98.93} $\pm$ \textbf{0.13} & 97.48 $\pm$ 10.56 & 99.94 $\pm$ 0.04 \\
bag & 71.03 $\pm$ 10.88  & \textbf{92.28} $\pm$ \textbf{1.71} & 96.87 $\pm$ 0.17 & 98.86 $\pm$ 0.12 \\
ankle boot & 90.50 $\pm$ 11.06  & \textbf{99.22} $\pm$ \textbf{0.23} & 97.39 $\pm$ 1.33 & 99.92 $\pm$ 0.12 \\ \hline
\end{tabular}
\label{fashion}
\end{table}
\endgroup

\subsubsection{CIFAR-10}
The CIFAR-10 dataset consists of 60,000 images in the $32 \times 32 \times$ RGB format. It includes 10 classes: airplane, automobile, bird, cat, deer, dog, frog, horse, ship, and truck.
We chose airplane as the positive class and preprocessed the original dataset in the same way as the experiment with the Fashion-MNIST dataset.

In this experiment, we used a convolutional neural network model with the following architecture:
\vspace{-0.3\baselineskip}
\begin{itemize}
  \setlength{\itemsep}{0pt}
  \item Convolution (3 in- /18 out-channels, kernel size 5).
  \item Max-pooling (kernel size 2, stride 2).
  \item Convolution (18 in- /48 out-channels, kernel size 5).
  \item Max-pooling (kernel size 2, stride 2).
  \item Fully-connected (800 units) with ReLU.
  \item Fully-connected (400 units) with ReLU.
  \item Fully-connected (1 unit).
\end{itemize}
\vspace{-0.3\baselineskip}

The results of the CIFAR-10 experiment are shown in Table \ref{cifar}.
Adjusted Pconf outperformed original Pconf in all cases.
Although the accuracy of adjusted Pconf was significantly lower than that of the fully-supervised method, adjusted Pconf was comparable to original Pconf with non-skewed confidence in some cases.

\begingroup
\renewcommand{\arraystretch}{1.3}
\begin{table}[!t]
\centering
\caption{The mean and standard deviation of the classification accuracy over 20 trials using convolutional neural network models on the CIFAR-10 dataset. Adjusted Pconf classification was compared with original Pconf classification, original Pconf classification with non-skewed confidence, and fully-supervised classification. We used the airplane as the positive class in all cases. Bold face denotes the best and comparable methods according to the paired $t$-test at the significance level of 5\% between original Pconf and adjusted Pconf.}
\begin{tabular}{l|ll|l|l} \hline
Negative & \multicolumn{1}{c}{O. Pconf} & \multicolumn{1}{c}{A. Pconf} & \multicolumn{1}{|c}{Non-skewed} & \multicolumn{1}{|c}{Supervised}  \\ \hline
automobile & 68.36 $\pm$ 6.27  & \textbf{79.66} $\pm$ \textbf{2.71} & 82.68 $\pm$ 0.76 & 93.35 $\pm$ 0.48 \\
bird & 68.42 $\pm$ 4.12  & \textbf{80.78} $\pm$ \textbf{1.67} & 82.29 $\pm$ 1.15 & 88.73 $\pm$ 0.66 \\
cat & 76.44 $\pm$ 7.16  & \textbf{85.08} $\pm$ \textbf{2.86} & 86.57 $\pm$ 1.02 & 92.35 $\pm$ 0.37 \\
deer & 74.16 $\pm$ 8.69  & \textbf{84.87} $\pm$ \textbf{2.30} & 86.71 $\pm$ 0.50 & 92.95 $\pm$ 0.52 \\
dog & 77.07 $\pm$ 6.24  & \textbf{85.28} $\pm$ \textbf{4.19} & 88.97 $\pm$ 1.51 & 94.24 $\pm$ 0.30 \\
frog & 83.62 $\pm$ 4.16  & \textbf{89.07} $\pm$ \textbf{1.28} & 90.34 $\pm$ 0.79 & 94.98 $\pm$ 0.38 \\
horse & 67.03 $\pm$ 9.75  & \textbf{77.02} $\pm$ \textbf{8.29} & 86.18 $\pm$ 1.02 & 94.68 $\pm$ 0.35 \\
ship & 57.55 $\pm$ 3.74  & \textbf{69.00} $\pm$ \textbf{1.98} & 69.87 $\pm$ 1.45 & 88.43 $\pm$ 0.50 \\
truck & 70.11 $\pm$ 4.15  & \textbf{81.25} $\pm$ \textbf{1.25} & 83.01 $\pm$ 0.88 & 90.85 $\pm$ 0.44 \\ \hline
\end{tabular}
\label{cifar}
\end{table}
\endgroup

\section{Real-World Application}

In this section, we apply original and adjusted Pconf classification to the real-world problem of drivers' drowsiness prediction.
Through the experiment, we show that the confidence may be skewed in a real scenario, and can be corrected by our proposed method.

The objective of this application is to predict drivers' drowsiness from the heartbeat information. 
Typically, a classifier for this problem can be trained with an experimental dataset including both \emph{Alert} and \emph{Drowsy} states collected by using a driving simulator.
However, there are two issues with this way of building a classifier.
The first one is derived from the fact that a driver reacts differently to a driving simulator and the real road situation \cite{hallvig2013sleepy}.
A classifier trained with data from a driving simulator experiment may be less useful in the real traffic environment, but an experiment in the real environment is difficult and dangerous because drowsy driving easily leads to accidents.
The second one is that driving until feeling strong drowsiness and the manual annotation of drivers' states are extremely time-consuming even though we use a driving simulator.
For these two issues, there is a strong motivation to learn a binary classifier without collecting \emph{Drowsy} samples.

\subsection{Drivers' Drowsiness Dataset}
We used an in-house drivers' drowsiness dataset \cite{Kaji2018}, which was constructed from the driving simulator experiment.
Three healthy males engaged in the driving task with a driving simulator along an expressway around 100 km/h overtaking other cars if necessary.
The task continued until the strong drowsiness was observed or a driver finished the whole course (about 150 km).
The facial expressions and the electrocardiogram (ECG) signal were recorded during the driving task to construct the dataset which includes the feature vector $\bm{x}$ and the drowsiness score $D$.
The feature vector $\bm{x}$ was composed of seven heart-rate-variability indicators computed in the frequency domain (the spectral power of the low frequency component, etc) and time domain (the variance of the peak-to-peak interval of the R-wave, which is the largest wave of ECG, etc.).
These features were computed at 60 seconds intervals with 120-second sliding windows, and normalized to have zero mean and unit standard deviation for each driver.
Expert $j$ ($j=1,2,3$) independently evaluated the drowsiness score $D_{j}$ , which was rated from 1 (``Not sleepy'') to 5 (``Very sleepy''), 
based on the recorded facial expressions \cite{Kitajima1997} every 60 seconds.

Each driver performed the driving tasks 10 times, and only five trials were annotated and the remaining five trials were unlabeled.
We used the annotated five trials in this study.

\subsection{Experimental Setup}
We divided the collected data into \emph{Alert} (positive) and \emph{Drowsy} (negative) classes to reframe the drowsiness prediction as a binary classification problem.
The samples with the median of the drowsiness score $D_{j}$ less than 3 were labeled as \emph{Alert}, otherwise the samples were labeled as \emph{Drowsy}.
We also calculated the positive confidence from the drowsiness score $D_{j}$ as follows, which is confined between 0 and 1:
\[
r =1- \frac{1}{12}\sum_{j=1}^{3} \left(D_{j}-1\right).
\]

We compared the performance of original Pconf classification, adjusted Pconf classification, and fully-supervised classification.
The logistic regression with the Gaussian kernel of bandwidth 1.0 was used for all of these methods.

In this experiment, the evaluation of the classification performance was conducted independently for each driver.
More specifically, the samples of four driving trials by the same driver were used for training a classifier, and the samples of the remaining one trial were used as a test set.
We repeated this procedure five times for each driver so that all trials were tested once.
The F-measure was used as the performance metric since the dataset was imbalanced.

We computed the mean FN rate over 20 trials with fully-supervised classification for each driver, and then used the average of the mean FN rate of two drivers as an estimate of $\phi$ for the remaining one driver.
Only \emph{Alert} samples and their confidence were used for training original Pconf and adjusted Pconf classifiers, and hyperparameter tuning.

\subsection{Results and Discussion}
The results are reported in Table \ref{ds}.
The F-measure could not be calculated for original Pconf because it predicted all the samples to be \emph{Alert}.
This result suggests that there existed strong skew towards the \emph{Alert} class in the confidence.
It can be explained by the fact that it is difficult even for the experts to find out the sign of drowsiness on driver's face because how the facial expressions change when she gets sleepy varies across drivers.
Consequently, the drowsiness score tends to be lower than the actual drowsiness of drivers.
Such bias can arise in any situation where there is asymmetric difficulty in evaluating confidence.

On the other hand, adjusted Pconf worked reasonably well with the drowsiness dataset.
This result indicates that our method is also effective for the real-world problem where the positive confidence is given based on the subjective rating.

\begingroup
\renewcommand{\arraystretch}{1.2}
\begin{table}[!t]
\centering
\caption{The mean and standard deviation of the F-measure over 20 trials using kernel models on the drivers' drowsiness dataset. Adjusted Pconf classification was compared with original Pconf classification, and fully-supervised classification. N/A indicates that the F-measure could not be calculated since the output was \emph{Alert} (positive) for all the test samples.}
\begin{tabular}{c|cc|c} \hline
Driver & O. Pconf & A. Pconf & Supervised  \\ \hline
1 & N/A  & 68.37 $\pm$ 0.96 & 59.92 $\pm$ 2.49 \\
2 & N/A  & 48.86 $\pm$ 0.50 & 66.29 $\pm$ 1.68 \\
3 & N/A  & 50.86 $\pm$ 0.87 & 55.66 $\pm$ 5.33 \\ \hline
\end{tabular}
\label{ds}
\end{table}
\endgroup

\section{Conclusion}

In this paper, we proposed adjusted Pconf classification to mitigate the influence of skewed confidence, which is not easy to handle but naturally emerges in the real-world applications.
The key idea is the assumption of the misclassification rate of positive samples, which is minimal statistical information as a prior knowledge.
The hyperparameter to correct skew in confidence is optimized through the minimization of the squared difference between the misclassification rate and validation error.
We demonstrated the effectiveness of the proposed method through the synthetic and benchmark experiments, and the real-world drowsiness prediction problem.

\bibliographystyle{named}
\bibliography{ijcai20}

\end{document}